\DocumentMetadata{
      pdfversion=2.0
    }

\documentclass[sigconf,natbib=false]{acmart}

\AtBeginDocument{%
  }

\setcopyright{cc}
\setcctype{by-nc-nd}
\acmDOI{10.1145/3757279.3788671}
\acmYear{2026}
\copyrightyear{2026}
\acmISBN{979-8-4007-2128-1/2026/03}
\acmConference[HRI '26]{Proceedings of the 21st ACM/IEEE International Conference on Human-Robot Interaction}{March 16--19, 2026}{Edinburgh, Scotland, UK}
\acmBooktitle{Proceedings of the 21st ACM/IEEE International Conference on Human-Robot Interaction (HRI '26), March 16--19, 2026, Edinburgh, Scotland, UK}
\received{2025-09-30}
\received[accepted]{2025-12-23}

\RequirePackage[
  datamodel=acmdatamodel,
  style=acmnumeric,
  sortcites=true,  
  ]{biblatex}
\usepackage{algorithm}
\usepackage[noend]{algpseudocode}
\usepackage{multirow}
\usepackage{wrapfig}
\usepackage[table]{xcolor}
\usepackage{xurl}
\usepackage{fontawesome}

\begin{document}

\title{Interface-Aware Trajectory Reconstruction of Limited Demonstrations for Robot Learning}

\author{Demiana R. Barsoum}
\affiliation{%
  \institution{Northwestern University \& Shirley Ryan AbilityLab}
  \city{Chicago}
  \state{Illinois}
  \country{USA}}
\orcid{0009-0001-2831-1615}
\email{dbarsoum@u.northwestern.edu}

\author{Mahdieh Nejati Javaremi}
\affiliation{%
  \institution{Northwestern University \& Shirley Ryan AbilityLab}
  \city{Chicago}
  \state{Illinois}
  \country{USA}}
\orcid{0000-0003-4368-4572}
\email{m.nejati@u.northwestern.edu }

\author{Larisa Y.C. Loke}
\affiliation{%
  \institution{Northwestern University \& Shirley Ryan AbilityLab}
  \city{Chicago}
  \state{Illinois}
  \country{USA}}
\orcid{0009-0000-7015-2649}
\email{larisaycl@u.northwestern.edu}

\author{Brenna D. Argall}
\affiliation{%
  \institution{Northwestern University \& Shirley Ryan AbilityLab}
  \city{Chicago}
  \state{Illinois}
  \country{USA}}
\orcid{0000-0002-4280-8492}
\email{brenna.argall@northwestern.edu}

\begin{abstract}
Assistive robots offer agency to humans with severe motor impairments. Often, these users control high-DoF robots through low-dimensional interfaces---such as using a 1-D sip/puff interface to operate a 6-DoF robotic arm. This mismatch results in having access to only a \textit{subset} of control dimensions at a given time, imposing \textit{unintended and artificial} constraints on robot motion.
As a result, interface-limited demonstrations embed suboptimal motions that reflect interface restrictions rather than user intent. To address this, we present a trajectory reconstruction algorithm that reasons about task, environment, and interface constraints to \textit{lift} demonstrations into the robot’s full control space. We evaluate our approach using real-world demonstrations of ADL-inspired tasks performed via a 2-D joystick and 1-D sip/puff control interface, teleoperating two distinct 7-DoF robotic arms. Analyses of the reconstructed demonstrations and derived control policies show that lifted trajectories are faster and more efficient than their interface-constrained counterparts while respecting user preferences.
\end{abstract}


\maketitle

\section{Introduction}

Across many teleoperation settings, a persistent challenge arises when the dimensionality of the control interface does not match that of the robot. This mismatch is especially pronounced for people with motor impairments operating high degree-of-freedom (DoF) assistive devices~\cite{hagengruber2025assistive}. Teleoperation efficacy is predicated on the interface~\cite{erdogan2017effect}, yet severe motor impairments often limit users to low-dimensional devices such as switch-based head arrays or 1-D sip/puff interfaces~\cite{fehr2000clinical}. Consequently, the user's available control actions span only a subset of the robot’s operating space, often leading to fatigue and increased cognitive and physical burden~\cite{herlant2016assistive}.


One strategy to ease this burden is to incorporate assistive autonomy into the teleoperation loop~\cite{li2023classification}. When users of physically assistive devices prefer to remain in the control loop, as they so often do, it becomes important to position them not only as operators but also as \textit{teachers} who guide the robot’s autonomous behavior. A widely adopted approach for such instruction within robotics is Learning from Demonstration (LfD)---which is inspired by how humans teach and learn from one another. While traditional LfD typically assumes demonstrators with full mobility, we contend that if assistive robots are to serve people with motor impairments, then it must be feasible for the end-users themselves to be the teachers. 



Demonstrations collected through interfaces accessible to those with severe motor impairments typically are inherently limited in dimensionality.
Crucially, the assumption that these demonstrations fully capture the user’s intended motion thus becomes invalid.
We therefore propose that robot learning systems should infer and \textit{reconstruct} the richer task-space behavior intended by the human, rather than directly replicate interface-limited demonstrations. Our key insight is that \textit{while teleoperation is restricted by the interface, autonomous control is not}. Instead, when appropriate, autonomy should leverage the full, unrestricted robot control space.


In this work, we introduce an approach that reconstructs trajectories from interface-constrained demonstrations by reasoning about interface, task, and environment constraints in order to lift demonstrations into the robot’s full control space. These reconstructed trajectories then can be used to train control policies, deployed within fully autonomous or shared-control paradigms. Although motivated by assistive robotics, this capability is broadly useful:
\textit{any} robotics application with teleoperation interfaces that struggle to match the higher dimensionality of the robot might benefit.
In the case of assistive robotics, it opens the door to empowering a new class of robot teachers: persons for whom the combination of motor impairment and accessible interface limits the channels through which they can provide information for robot instruction.  

The contributions of this work are the following:
\begin{enumerate}

    \item We present an interface-aware, multi-dimensional trajectory reconstruction algorithm that lifts interface-constrained demonstrations into the robot’s unconstrained control space.


    \item We collect demonstrations on two distinct 7-DoF robotic arms using two control interfaces \textit{accessible} to persons with severe motor impairments that are industry standards for power wheelchair operation, including the extremely limited 1-D sip/puff and a 2-D joystick, and evaluate our algorithm across six diverse, real-hardware ADL-inspired tasks.


    \item We demonstrate that autonomous policies trained on reconstructed (lifted) demonstrations achieve improved performance using the same model complexity, demonstrating the utility of our method for robot learning.
\end{enumerate}

The rest of the paper is organized as follows. We discuss related work in Section~\ref{sec:background}, followed by our framework and technical contribution in Section~\ref{sec:method}. Our experimental setup is presented in Section~\ref{sec:experiments}, followed by results in Section~\ref{sec:results}. Limitations and future work are discussed in Section~\ref{sec:limitations}, and conclusions in Section~\ref{sec:conclusion}. 

\section{Related Work} \label{sec:background}

\textbf{Learning from \textit{Limited} Demonstrations.} In LfD, robots learn task policies from example trajectories~\cite{ARGALL2009469}. Demonstrations may be obtained through kinesthetic teaching~\cite{gen_lfd2009,calinon2007}, video-based imitation~\cite{finn2017humanvideo, smith2019humanvideo, bahl2022humanvideo, qin2022humanvideo, wang2023mimicplayhumanvideo}, commercial virtual reality (VR) controllers~\cite{arunachalam2023holoVR, shridhar2023perceiverVR}, inertial measurement unit (IMU) sensors~\cite{kim2009walkingimu, laghi2018sharedimu}, or exoskeleton-based
teleoperation~\cite{wu2023gello, yang2024ace}, among others. Most LfD methods assume the demonstrators are experts and the collected trajectories are optimal or near-optimal~\cite{IL_survey2017,pmlr-v15-ross11a}.
      

Prior work on \textit{limited} demonstrations typically focuses on having few demonstrations~\cite{duan2017oneshot} or on relaxing expert-quality assumptions~\cite{NIPS2013_lfld}. In contrast, our interest lies in \textit{dimensionally-limited} demonstrations---cases where the control interface offers fewer degrees of freedom than the robot’s task space. Here, the demonstrations are constrained not by expertise or sample count but by the interface itself. This setting introduces unique challenges for LfD and motivates the need for reconstructing higher-dimensional trajectories that reflect the user’s intended task-space behavior.

\noindent\textbf{Dimension Mismatch.} Automated mode switching is one approach to reduce the extra cognitive burden~\cite{herlant2016assistive,gopinath2017mode} of control mode switches, although it requires feedback to the human to maintain transparency. The appropriate method of feedback is an active field of research~\cite{bolano2018transparent}. Another method is to change the mapping of what the human is commanding (essentially dynamic mode shaping)~\cite{mehta2021learning}, which removes manual mode switching but limits the user’s control over mode selection. Alternatively, a substantial line of research learns a latent action space that enables low-dimensional interface control of high-DoF robots~\cite{losey2021learning,2021LILA}. These methods typically rely on kinesthetic demonstrations from unimpaired operators and, using \textit{dimensionality reduction techniques}, task-dependent lower-dimensional embeddings or motion primitives are mapped to the interface activations. Furthermore, the most limited interfaces in prior work still provide at least two simultaneous control signals.

In contrast, our setting involves demonstrations collected exclusively through interfaces accessible to people with severe motor impairments. Thus, kinesthetic teaching is not a viable option. Instead, what the user is able to provide are control signals issued through a constrained, low-dimensional interface. Rather than learning lower-dimensional task embeddings to match the interface, we address the inverse problem: mapping dimensionally-limited demonstrations into the \textit{higher-dimensional} control space of the robot.

\noindent\textbf{Correspondence Matching.} Our work is distinct from prior research on correspondence shift and correspondence matching~\cite{alissandrakis2007correspondence}. In our context, the embodiment of the robotic device is the same between the learner and the demonstrator; however, there is a mismatch in the sequencing and dimensionality of the actions. The demonstrator's actions lie on a lower-dimensional manifold compared to the actions available to the robotic device. Although the robot \textit{could} learn to directly imitate the demonstrator's low-dimensional commands, our strategy is to
reason beyond the demonstrator's limited control input for richer policies.


\noindent\textbf{Constrained Policy Learning.} Constrained policy learning typically seeks to preserve and incorporate the constraints observed in demonstrations~\cite{armesto2017learning}. In our setting, however, many constraints arise not from the task but from the limitations of the user’s control interface. The agent must learn to \textit{avoid} reproducing interface-induced constraints to generalize beyond the spatiotemporal limitations of the demonstrations and improve task efficiency. Previous work incorporates interface-constraint rules in a heuristic graph-based pathfinding algorithm~\cite{Broad_Argall_2016}; in contrast, our approach learns behaviors directly from user-provided demonstrations.

Additionally, our motivation for seeking higher-dimensional trajectories is predicated on prior work demonstrating that increasing the number of accessible control dimensions enables the optimization of secondary criteria such as energy and jerk, leading to smoother trajectories~\cite{chembuly2018trajectory, dai2020planning}, which correlate with reduced task difficulty and improved task time and success~\cite{javaremi2020characterizationassistiverobotarm}. Higher-dimensional control also improves long-term task performance~\cite{rizzoglio2023non}. Together, these findings motivate removing interface constraints on demonstrator trajectories to utilize the robot’s full control space. 

\section{Methodology} \label{sec:method}
We present our framework that maps dimensionally-limited demonstrations to the higher-dimensional robot control space.

\subsection{Problem Formulation} 
We define the \textit{interface dimensionality} $l$ as the maximum number of control channels that an interface can activate simultaneously. Let the 
\textit{robot dimensionality} $m$ be the number of controllable degrees-of-freedom (DoFs) of the robot hardware. The robot control vector $\boldsymbol{u} \in \mathcal{K}$ is defined on a smooth, $m$-dimensional manifold $\mathcal{K}$. 

\begin{figure}[h]
  \begin{minipage}[c]{0.23\textwidth}
    \includegraphics[width=\textwidth]{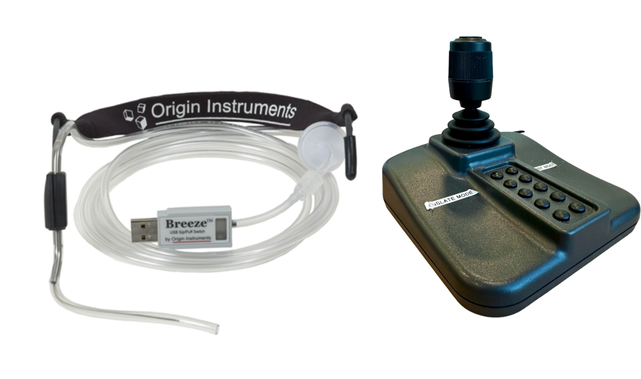}
  \end{minipage}%
  \hfill
    \begin{minipage}[c]{.8\textwidth}
    \small
    \begin{tabular}{c@{\hskip 6pt}|@{\hskip 6pt}c@{\hskip 6pt}c@{\hskip 6pt}c@{\hskip 6pt}c@{\hskip 6pt}c@{\hskip 6pt}c@{\hskip 6pt}c}
    \hline
    \textbf{Control Mode:} & \textbf{1} & \textbf{2} & \textbf{3} & \textbf{4} & \textbf{5} & \textbf{6} & \textbf{7} \\
    \hline
    \textbf{Sip/Puff} & $v_x$ & $v_y$ & $v_z$ & $\omega_x$ & $\omega_y$ & $w_z$ & g \\
    \textbf{Joystick} & $v_x$, $v_y$ & $v_z$, $\omega_z$ & $\omega_x$, $\omega_y$ & g & --- & --- & --- \\
    \hline
    \end{tabular}
    \end{minipage}
  \caption{Limited interfaces and control modes. Sip/Puff Breeze\textsuperscript{\texttrademark} with Headset (\textit{left}). Image \copyright~ Origin Instruments~\cite{OriginInstruments}, reproduced with permission. 2-D joystick (\textit{right}). An example of modal partitioning of the robot control space (\textit{bottom}, end-effector translation $\vec{v}$ and rotation $\vec{\omega}$ velocity vectors, and gripper $g$ open/close.)}
  \Description{The left image showcases the two control interfaces used to collect demonstrations: a sip/puff and a joystick. The left is a table depicting the paradigms for each of the control interfaces. the sip/puff requires 7 modes to control each dimension plus the gripper control while the 2-D joystick offers four modes that give access to 2 dimensions at a time.}
  \label{fig:interfaces}
  \vspace{-4mm}
\end{figure}

\noindent\textbf{Interface Constraints.}
Within the field of robot teleoperation, often the dimensionality of the human control interface does not fully span the dimensionality of the robot control space, and so $l < m$. This is especially pronounced for very high-dimensional robots, as well as for the realm of assistive robots teleoperated by low-dimensional interfaces (e.g., Figure~\ref{fig:interfaces}). 

The dimensionality mismatch between interface and robot hardware necessitates segmenting the control space into \textbf{control modes} that each span a subset of the dimensions of $\boldsymbol{u}$. Let $\mathcal{M}$ be the set of all control modes $\boldsymbol{u_i}$, where $dim(\boldsymbol{u_i}) = n < m$ and the union of all control modes covers the robot control manifold, $\bigcup_{i=1}^{|\mathcal{M}|} \boldsymbol{u_i} = \mathcal{K}$.

The added complexity of switching between control modes during teleoperation increases the mental workload and the operational burden on users~\cite{herlant2016assistive}, not only restricting the \textit{quantity} of demonstrations that can be practically collected but also restricting how the user operates the robot and therefore limiting the \textit{quality} and \textit{range} of demonstrations that can be collected. These limitations inform our algorithmic framework described in Section~\ref{sec:algo}.

\subsection{Interface-Aware Trajectory Reconstruction}\label{sec:algo}
We present a method for multi-dimensional, interface-aware trajectory reconstruction of demonstrations collected using lower-dimensional interfaces.
In brief, our method first segments a trajectory according to control mode (Algorithm~\ref{alg:split_traj}), then reasons about interface, task, and environment constraints (Algorithms~\ref{alg:constraints} \&~\ref{alg:constraints_check}), and finally 
lifts trajectories into higher dimensions via a composition algorithm (Algorithm~\ref{alg:reconstruct}).
Details for each algorithm are presented in Sections~\ref{sec:segment}-\ref{sec:reconstruct}.

We first define a demonstration trajectory $D$ to consist of tempor\-ally-sequenced tuples: $D = \{ (\boldsymbol{q}^t,\boldsymbol{\mu}^t,\boldsymbol{x}^t )\}_{t=0}^T $,
where $\boldsymbol{q}^t \in \mathbb{R}^{N_q}$ is the robot state at time $t$, $\boldsymbol{\mu}^t \in \mathbb{R}^{N_{\mu}}$ is the control mode mask, and $\boldsymbol{x}^t \in \mathbb{R}^{N_x}$ is the world state relevant to the task. Note that the robot state $\boldsymbol{q}^t$ might be defined in task space or joint space, and consist of position, velocity, or both (as relevant to the demonstrated task). The control mode mask is a vector of length $m$ that provides a binary indication of which control dimensions are able to be operated at timestep $t$, as dictated by the current control mode.\footnote{For the sip/puff, the mask is one-hot as the demonstrator only has access to one control dimension per mode. For the 2-D joystick, it can be one- or two-hot.}

\subsubsection{Trajectory Segmentation By Mode\label{sec:segment}}
Our method first segments the demonstration trajectory $D$ according to control mode switches, in Algorithm~\ref{alg:split_traj}. The aim is to construct a set $\mathbb{S}$ of trajectory segments $S$, where $S$ is a temporally-contiguous block of trajectory points for which the same control mode was operational.

The algorithm first filters out transient mode switches that result from mode cycling (lines 5-6).\footnote{Mode selection with a limited interface often is linear or cyclical, meaning that one needs to cycle through undesired modes on the way to selecting the intended mode. For paradigms that allow for direct mode selection without cycling (e.g., via button press), this filtering step can be omitted.}  
Then, if a mode switch has not occurred, the trajectory point $D^t[\cdot]=(\boldsymbol{q}^t,\boldsymbol{\mu}^t, \boldsymbol{x}^t)$ is added to the current segment $S$ (lines 8-9); otherwise, if a mode-switch has occurred, a new segment is started (line 13). We additionally filter out artifacts from spurious unintended mode selections (lines 11-12). 
Here, $\epsilon$ is a threshold on minimum trajectory length (in our implementation, $\epsilon$ = 100, $\sim1s$).

\begin{algorithm}[h]
\scriptsize
    \caption{Segment Trajectory By Mode}\label{alg:split_traj}
    \begin{algorithmic}[1]
        \Function{SegmentByMode}{$D$} \Comment{Segments the path based on mode switch}
            \State $\mathbb{S} \gets [\,] $  
            \State $S \gets D^{t=0}[\cdot]$     
            \For {$t \gets 1$ \textbf{to} $T$}
                \If{$\boldsymbol{\mu}^{t-1} \neq \boldsymbol{\mu}^t \neq \boldsymbol{\mu}^{t+1}$}
                    \State Discard datapoint. \Comment{Cycling through modes during mode switch}
                \Else
                    \If{$\boldsymbol{\mu}^{t-1} = \boldsymbol{\mu}^t$} \Comment{If the modes are the same}
                        \State $S \gets [S; D^t[\cdot]]$ 
                    \Else \Comment{Else the modes are different}
                        \If{ $\epsilon \leq$ length($S$)}  \Comment{Filter out mode switch artifacts}
                            \State $\mathbb{S} \gets \{\mathbb{S}, S\}$ 
                        \EndIf
                        \State Initialize $S \gets D^t[\cdot] $
                    \EndIf
                \EndIf
            \EndFor
            \State \Return $\mathbb{S}$ 
        \EndFunction
    \end{algorithmic}
\end{algorithm}

\subsubsection{Apply Constraints to Reconstruction}
Once we have segment\-ed the trajectory by mode, we reason about interface, task, and environment constraints to determine whether two segments should be composed.
This is achieved in Algorithm~\ref{alg:constraints} by performing a recursive pairwise comparison of consecutive trajectory segments $S_i$ and $S_{i+1}$  (lines 6-11), merging them if they meet the requisite constraint criteria (line 12), and updating the full trajectory with the higher dimensional merged segment $S_{i^*}$ (line 13). 
The process is repeated (line 18) until no valid pairwise mergers remain. 

\begin{algorithm}[h]
\scriptsize
\caption{Apply Constraints}\label{alg:constraints}
\begin{algorithmic}[1]
\Function{ApplyConstraints}{$~\mathbb{S}~$}
    \newcommand{\NOT}{\textbf{not }}
    \newcommand{\AND}{\textbf{and }}
    \newcommand{\OR}{\textbf{or }}
    \newcommand{\cont}{\textbf{continue }}
    \State $\Delta \gets \mathtt{False}$
    \State $N_s \gets |\mathbb{S}|$
    \State $i \gets 0$ 
    \While{$i < N_s-1$}
    
            \If{ \Call{Environ\_Constraints}{$S_i$} \OR \Call{Environ\_Constraints} {$S_{i+1}$}} \label{ln:env_check_start} \Comment{Algorithm~\ref{alg:constraints_check}} 
            \State $i \gets i+1$ 
            \State \cont
            \EndIf
            
            \If{ \Call{Task\_Constraints}{$S_i$} \OR\Call{Task\_Constraints}{$S_{i+1}$}} \Comment{Algorithm~\ref{alg:constraints_check}} 
            \State $i \gets i+1$ 
            \State \cont 
            \EndIf \label{ln:task_check_end}


                    \State{$S_{i^*} \gets$ \Call{ReconstructSegments}{$S_i, S_{i+1}$} \Comment{Algorithm~\ref{alg:reconstruct}}} 
                    \State{$S \gets \lbrace S_{[0:i-1]}, ~S_{i^*}, ~S_{[i+2:N_s]} \rbrace$} \Comment{Replace segments $S_i$ and $S_{i+1}$ with $S_{i^*}$}
                    \State $\Delta \gets \mathtt{True}$
                    \State $N_s \gets |\mathbb{S}|$
        \State{$i \gets i+1$}
    \EndWhile
    \If {$\Delta = \mathtt{True}$}
        \State{ $ \mathbb{S} \gets$ \Call{ApplyConstraints}{$\mathbb{S}$}} \Comment{Iterate through segments again}
    \Else
        \State \Return $\mathbb{S}$
    \EndIf
\EndFunction
\end{algorithmic}
\end{algorithm}

In detail, we identify three forms of constraints to which segments of a trajectory can be subjected. \textbf{Environmental Constraints} and \textbf{Task Constraints} (lines 6 and 9) constrain the robot in response to environmental factors such as safety, and factors that are necessary for task completion. These constraints should be and are preserved.
By contrast, \textbf{Interface Constraints} 
impose artificial constraints on the robot's motion due to limitations of the interface, as discussed above. These are the constraints we remove, if possible, by composing adjacent segments and thus lifting their dimensionality.



\subsubsection{Constraints Implementation Details} \label{app:constraints}
The \textbf{Task} and \textbf{Environment Constraint Checks} are specific to the particular 
task and operational environment. Here we provide implementation details on the definition of environment and task constraints that are specific to our experimental setup (detailed further in Section~\ref{sec:exp}), as well as how these constraints are checked (Algorithm~\ref{alg:constraints_check}). 

In our implementation, the Environment Constraint considers collision safety, and a segment is flagged as constrained if at any point the nearest obstacle distance (encoded in $\boldsymbol{x}^t$) falls below a threshold ($\delta$).
The Task Constraint considers gripper operation, and a segment during which the gripper is opening or closing is not lifted. This is to ensure that segments with conflicting gripper states (before/after an open/close operation) are not composed.

Segments that have not been flagged as having either Environment or Task Constraints are assumed to be interface constrained, 
and thus able to be composed using the subroutine described (next) in Algorithm~\ref{alg:reconstruct} and Section~\ref{sec:reconstruct}.
\begin{algorithm}[]
    \scriptsize
    \caption{Check Constraints}\label{alg:constraints_check}
    \begin{algorithmic}[1]
    \Function{Environ\_Constraints}{$~S~$}
        \State{ $C \gets \mathtt{False}$ }
        \State{ $\boldsymbol{x} \gets \mathtt{extract}( S )$ } \Comment{Extract world state vector from segment}
        \For{ $ i \gets 0 ~\textbf{to}~ \mathtt{length}(\boldsymbol{x})-1$ }
            \If{ $\boldsymbol{x}[i] < \delta $}
                \State $ C \gets  \mathtt{True} $ \Comment{Environment Constraint: End-effector came close to obstacle}
            \EndIf
        \EndFor
        \State \Return $C$
    \EndFunction
    \\
    \Function{Task\_Constraints}{$~S~$}
        \State{ $C \gets \mathtt{False} $}
        \State $\boldsymbol{q}_g \gets \mathtt{extract}( S )$ \Comment{Extract gripper state vector from segment}
        \For{ $i \gets 0 ~\textbf{to}~ \mathtt{length}(\boldsymbol{q}_g)-1$}
            \If{ $\boldsymbol{q}_g[i] \neq \boldsymbol{q}_g[i+1] $}
                \State $C \gets \mathtt{True}$ \Comment{Task Constraint: Gripper is opening/closing}
            \EndIf
        \EndFor
        \State \Return $C$
    \EndFunction
\end{algorithmic}
\end{algorithm}

\subsubsection{Reconstruct in Higher Dimensions\label{sec:reconstruct}}
Algorithm~\ref{alg:reconstruct} reconstructs a higher-dimensional trajectory segment by merging two segments together. Our algorithm temporally aligns the pairwise segments using \textbf{time warping} to be equal in length (lines 3-6).\footnote{Our implementation performs interpolation using an open-source Python software package Pulser~\cite{waveform2022pulser} that utilizes SciPy's interpolator module~\cite{ scipy2020interp}.} The temporally aligned segments then are reconstructed by first composing the robot state and control mask components (line 7, denoted as the $\textsc{Merge}$ operation) and after reconciling the environmental components 
(line 8, the $\textsc{Reconcile}$ operation). To perform these operations, we partition segment $S$ into components 
$S^{\mathtt{rob}}$, which contains the robot information $(\boldsymbol{q},\boldsymbol{\mu})$,
and $S^{\mathtt{env}}$, which contains the environment information ($\boldsymbol{x}$).

\begin{algorithm}[h]
\scriptsize
\caption{Reconstruct Segments}\label{alg:reconstruct}
\begin{algorithmic}[1]
\Function{ReconstructSegments}{$~S_1, S_2~$}
    \State  $N_1 \gets \mathtt{n\_rows} ( S_1 ), \quad N_2 \gets \mathtt{n\_rows} ( S_2 )$
    \If{$N_1 \leq N_2$}
        \State $S_1 \gets \Call{TimeWarp}{S_1}$ \Comment{Match temporal length}
    \Else
        \State $S_2 \gets \Call{TimeWarp}{S_2}$
    \EndIf
    \State $S_{i^*}^{\mathtt{rob}} \gets$ \Call{Merge}{$S_1^{\mathtt{rob}},S_2^{\mathtt{rob}}$}  \Comment{Compose robot components}
    \State $S_{i^*}^{\mathtt{env}} \gets$ \Call{Reconcile}{$S_1^{\mathtt{env}},S_2^{\mathtt{env}}$} \Comment{Reconcile environment}
    \State $S_{i^*} \gets [S_{i^*}^{\mathtt{rob}}, S_{i^*}^{\mathtt{env}}]$\\
    \Return $S_{i^*}$
\EndFunction
\end{algorithmic}
\end{algorithm}

The exact manner of environmental reconciliation depends on what is encoded in $\boldsymbol{x}$ and what the relevant Environmental Constraints are. In our implementation, Environmental Constraints pertain to safety 
and $\boldsymbol{x}$ monitors the distance to the nearest obstacle. The reconciliation thus is a \textbf{\textit{min}} operation over these distance values for two segment points being composed. 

\begin{figure}[h]
    \centering
    \includegraphics[width=.9\linewidth]{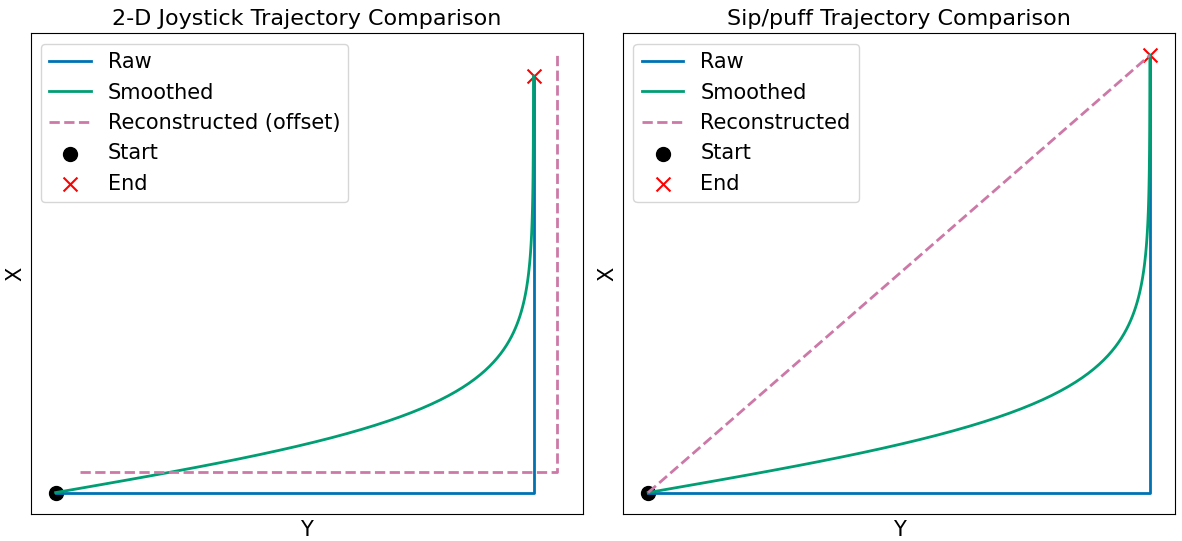}
    \caption{Illustrative example comparing raw, smoothed, and reconstructed trajectories from two different interfaces. \textit{Left:} The demonstrator intentionally issues lower-dimensional (1-D) commands through a higher-dimensional (2-D) interface. \textit{Right:} The demonstration instead is constrained by a lower-dimensional (1-D) interface.}
    \Description{Illustrative example of comparing raw, smoothed, and reconstructed trajectories. The figure on the right depicts an example when he demonstrator intentionally uses a higher-dimensional interface in a 1-D manner. Thus, the raw trajectory is a motion in the left, followed by motion up. The reconstructed trajectory matches the raw trajectory as this respects the user's intention and preference to use the interface in a 1-D manner. The figure on the right depicts an example of a 1-D interface (same right to up motion). However, the reconstructed trajectory couples the right and up motion since the user's preference cannot be embedded due to the nature of the interface. In both figures, the smoothed trajectory softens the trajectory, removing the user's preference/intention in the left figure.}
    \label{fig:2d_example}
    \vspace{-4mm}
\end{figure}

Note that any intentional temporal sequencing of dimension activations \textit{within} a trajectory segment (as illustrated in Figure~\ref{fig:2d_example}) will be preserved during reconstruction, as dimensions within a given segment are never composed with each other and time warping does not alter their temporal sequence.

\subsubsection{Illustrative Example}

We illustrate key features of our method in Figure~\ref{fig:2d_example}. These examples highlight how we preserve demonstrator preference when possible. Under the 2-D joystick (left), simultaneous X-Y control \textit{is} feasible, so the observed Y$\rightarrow$X sequence is interpreted to be intentional on the part of the demonstrator, and \textit{is not} lifted into a higher dimensional action space. In contrast, only sequential commands are possible with the 1-D sip/puff interface (right), so the same Y$\rightarrow$X sequence \textit{is} lifted into the full 2-D space.


In both cases, knowledge of the interface constraints dictates whether trajectory segments are composed or preserved. In comparison, traditional filtering methods (e.g., low-pass Butterworth filtering~\cite{butterworth1930theory}), smooth noise and soften local discontinuities but cannot reason about 
control interface limitations. Our algorithm does not yet account for when user preference is \textit{in alignment} with interface constraints; that is, if the Y$\rightarrow$X sequence was also intentional under sip/puff execution, and not an artifact of interface constraints. We leave this to future work.

\begin{figure}[h]
    \centering
    \includegraphics[scale=0.14]{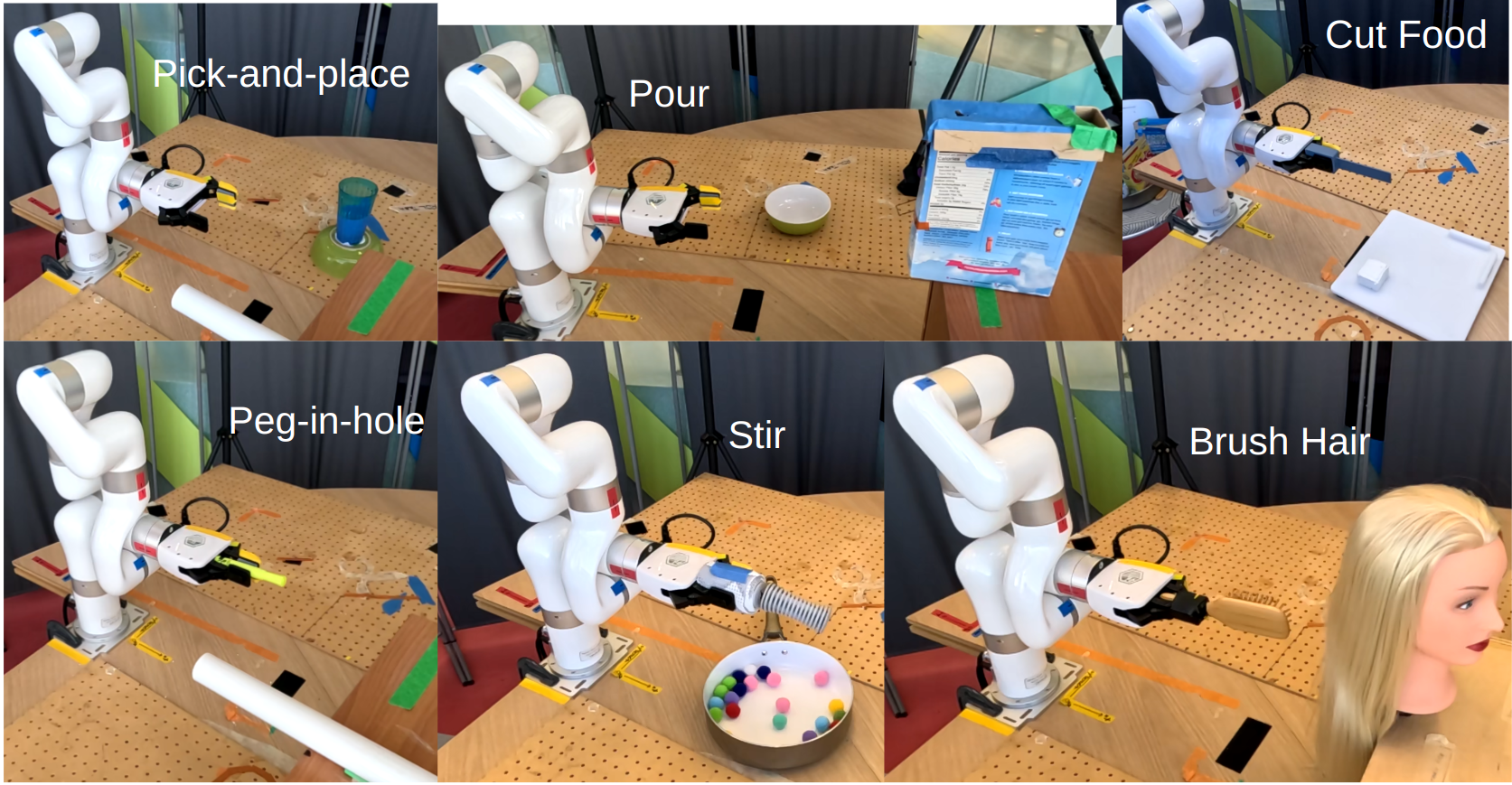}
    \caption{Experimental setup of each of the tasks. The xArm7 robotic arm with a pincher gripper is pictured in the same initial condition for each task.}
    \Description{Image of the 7-DoF robotic arm used for the experiments from UFactory, xArm7. The robot is white and includes a pincher gripper. The robot is in its initial configuration we name ``home". There are six number of tasks: Pick-and-place, Pour, Peg-in-hole, Stir, Brush Hair, and Cut Food.}
    \label{fig:hardware}
    \vspace{-5mm}
\end{figure}

\section{Experiments} \label{sec:experiments}
In this section, we describe our experimental setup and the real-hardware manipulation and ADL-like tasks~\cite{nanavati2023physically} conducted using two 7-DoF robotic arms. The goal of these experiments is to demonstrate the ability of our algorithm to effectively lift the limited demonstrations into a higher dimensionality, and to evaluate the lifted trajectories.

\subsection{Experimental Setup \label{sec:exp}}
\subsubsection{Hardware} 

We use a 7-DoF xArm7 robotic arm (Figure~\ref{fig:hardware}) from UFACTORY (Shenzhen, Guangdong, China), operated via ROS and the xArm SDK~\cite{xarmros, xarmsdk}, an open-source singularity-avoidance algorithm~\cite{guptasarma2025jparse}, and an adaptive tool frame for intuitive orientation control~\cite{vu2017adaptivetoolintuitive}. The arm is non-anthropomorphic, equipped with a pincher gripper, and supports six-dimensional end-effector control (3-D translation, 3-D rotation).


For teleoperation, we use two standard wheelchair-control interfaces (Figure~\ref{fig:interfaces}), both accessible to users with varying motor impairments. In particular, the \textit{3-axis joystick} (CH Products, California, USA) is actuated via deflection (most commonly enacted by the hand), and issues continuous-valued signals in 1-D or 2-D. Mode switching is accomplished via button press. The \textit{sip/puff} (Origin Instruments, Texas, USA) is actuated via respiration, and issues a continuous-valued 1-D (positive/negative) pressure signal, that for clinical applications typically is mapped to four discrete-valued commands via thresholding. In our implementation, two of these commands control positive/negative movement within a control dimension (`soft puff', `soft sip'), while the remaining two commands (`hard puff', `hard sip') control positive/negative cycling through modes to enact mode switches.

While the methodology is developed and evaluated primarily on the xArm7, we also assess its hardware translation on a Kinova Gen2 Jaco (Quebec, Canada) for a subset of tasks;
this second platform and results are included in Section~\ref{sec:jaco}.

\subsubsection{Tasks}
We evaluate the effectiveness of our algorithm on six ADL-inspired robot tasks. 
(1) Pick-and-place,
(2) Pour, 
(3) Peg-in-hole, 
(4) Stir, (5) Cut Food, and 
(6) Brush Hair.

\textbf{Pick-and-place.} Pick up a cup (placed right-side up) on one end of the workspace, transport it to the opposite end, and place it upside-down. This task is a proxy to unloading a dish rack.
\textbf{Pour.} Pick up a cereal box on a shelf and pour the cereal into a bowl on the table.
\textbf{Peg-in-hole.} Insert a Sharpie highlighter marker (peg) into a pipe (hole), placed directly in front of the robot on an elevated surface. The setup is designed to have minimal clearance (on the order of a few millimeters) to emulate realistic low-tolerance fit-insertion tasks.
\textbf{Stir.} Move the whisk to the bowl (placed directly in front of the robot), and stir one complete time.
\textbf{Cut Food.} Use a knife to cut tofu on a cutting board. The task is complete after two cuts.
\textbf{Brush Hair.} Complete one stroke of brushing from the top of the head to the ends of the hair.

For each of the tasks, the robot starts in the same initial configuration (Figure~\ref{fig:hardware}). In 4 out of 6 tasks (excluding Pick-and-place and Pour), the tool required to complete the task is already held in the robot gripper before data collection begins. To ensure reproducibility, we mark start and end positions in the environment. 

For each interface and task, our protocol consists of collecting demonstrations until 5 successful trials are observed, with a maximum of 20 attempted trials;
thus, difficult to teleoperate tasks can include as few as a single successful demonstration. The demonstrator teleoperating the robot is a researcher and an author.

We intentionally collect a small number of demonstrations in consideration of the target end-user population for this work. To teleoperate a 7-DoF non-anthropomorphic robotic arm using a 1-D interface is difficult to learn and exhibits high cognitive workload and fatigue. An expectation of many demonstrations under such constraints is unreasonable, and a learning algorithm that requires as much is unsuitable. Moreover, the complexity of performing these tasks with such low-dimensional interfaces underlines the need to automate these tasks. We thus aim to both (1) mimic the number of demonstrations we can reasonably ask a motor impaired teacher to provide and (2) test the efficacy of robot learning from very few demonstrations.


\subsection{Learning Formulation} 
A primary application of our algorithm is in policy learning. From a machine learning perspective, our algorithm serves as a data preprocessing step that transforms demonstrations constrained by low-dimensional control inputs into richer demonstration data. By processing the data through our algorithm first, we can then utilize any standard (supervised) machine learning framework to generate a policy mapping states to actions that utilizes the robot's available action space, rather than being restricted to a subset by the interface-constrained raw demonstrations. Our rule-based algorithm also eliminates the need for an expert human labeler (to determine task and environment constraints, or when segments might be lifted), allowing this step to be completed efficiently and automatically. 


We use behavior cloning to learn a policy from the demonstrations. Given the small dataset size---constrained by the need for feasibility among end-users with severe motor impairments---we implement a lightweight Multi-layer Perceptron (MLP) network using PyTorch~\cite{paszke2019pytorch} to learn each of the six tasks. 
We provide additional details about the network architectures and hyperparameters in Appendix~\ref{app:learning}. For tasks with greater than one collected demonstration, one demonstration is held out for testing and the training set is composed of the remaining demonstrations.

\section{Results} \label{sec:results}
We analyze the effectiveness of our interface-aware trajectory reconstruction algorithm in proof-of-concept experiments conducted on a physical robot.
We confirm successful lifting into higher dimensions, and compare the performance of our reconstructed trajectories against both the raw demonstrations and smoothed raw demonstrations (using the best performing filtering algorithm after tuning: Low-pass Butterworth Filter). 
For all tables, we indicate fewer than 5 collected demonstrations by gray-shading, and note the number of successful demonstrations collected in parenthesis.

\begin{figure}[b]
    \centering
    \includegraphics[width=.99\linewidth]{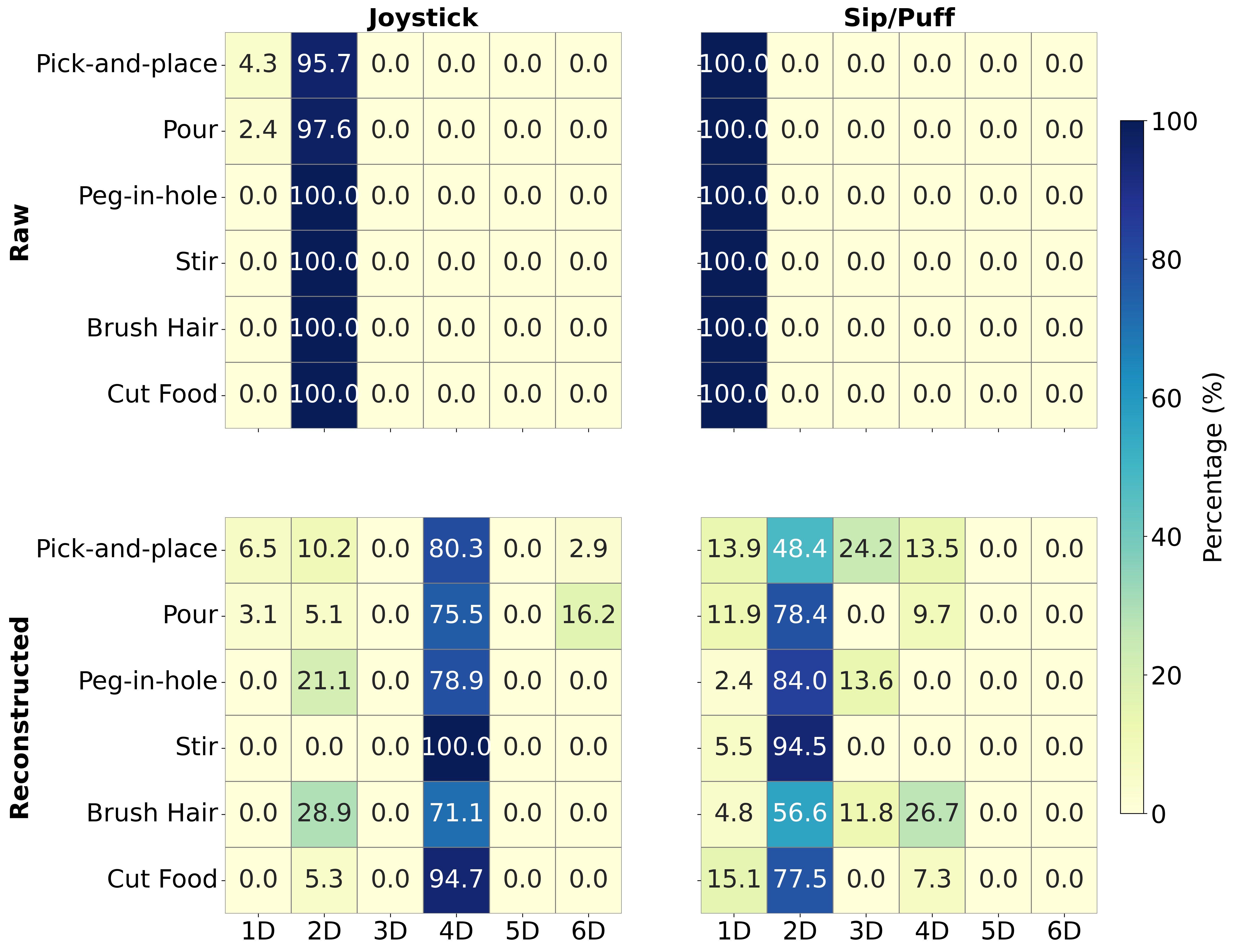}
    \caption{Percentage of dimensions activated across the full trajectory, averaged over all demonstrations. Raw (top row) and reconstructed (bottom row) trajectories collected with the joystick (left column) and sip/puff (right column.}
    \Description{This figure (for the xArm7 robotic arm) includes four confusion matrices (two-by-two). The first column of matrices is for the joystick control interface while the second column of matrices is for the sip/puff control interface. The top two matrices show the percentage of dimensions activated during the raw demonstrations while the bottom two matrices show the percentage of dimensions activated after our reconstruction (lifting) algorithm has been applied. Each row in the matrix represents one of the tasks. The joystick raw demonstrations (top left) go up to 2-D, while our reconstructed joystick demonstrations (bottom left) goes up to 6-D. Similarly, the sip/puff raw demonstrations (top right) are only one-dimensional, while with our method (bottom right), demonstrations are lifted to 4-D.}
    \label{fig:heatmap}
\end{figure}

\begin{figure}[b]
    \centering
    \includegraphics[width=0.75\linewidth]{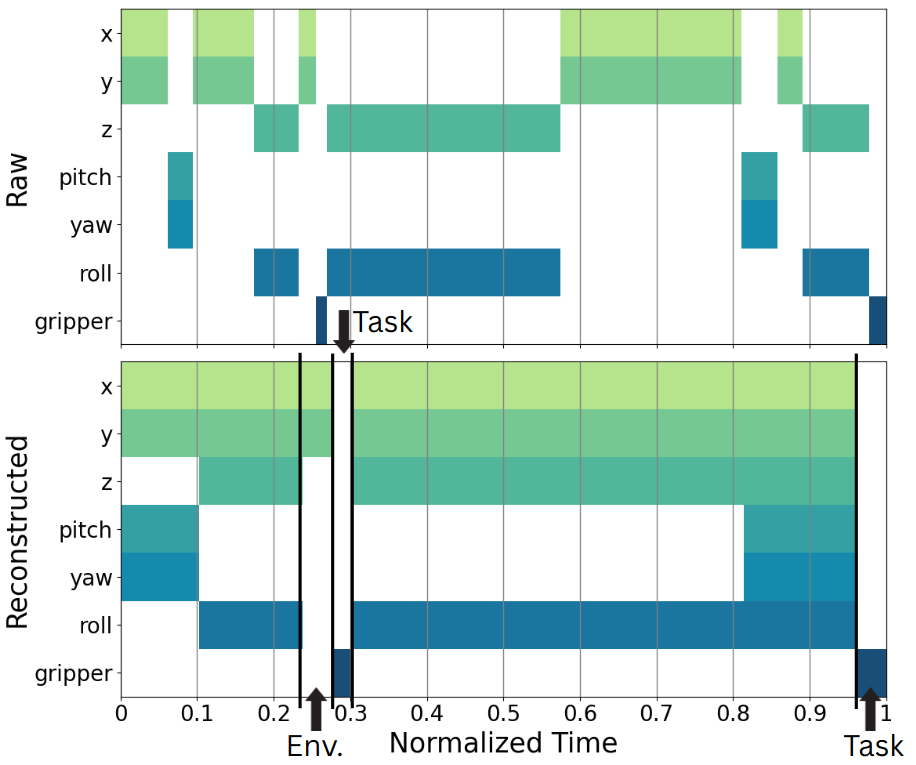}
    \caption{Evolution of dimension activations using the 2-D joystick for the Pick-and-place task. Activations in the raw demonstration (\textit{top}) are 1-D or 2-D only, while the reconstructed demonstration (\textit{bottom}) lifts as high as the full 6-D.}
    \Description{This figure is a timeseries plot depicting the evolution of dimension activation for a demonstration collection using the 2-D joystick joystick for the Pick-and-place task. The raw demonstration (top) shows 1-D to 2-D activation while the reconstructed demonstration (bottom) depicts the lifting into higher dimensions. In this case, up to the full 6-D. Environment and task constraints are demarcated in the timeseries.}
    \label{fig:dof_vs_time_plot}
    \vspace{-5mm}
\end{figure}

\subsection{Lifting to Higher Dimensions}
Figure~\ref{fig:heatmap} reports the number of control dimensions simultaneously activated for each timestep of a trajectory. Specifically, it reports the percentage of timesteps that achieve the simultaneous control of 1-6 dimensions, across all trajectories for a given task.
As expected, the raw demonstrations are capped at 1-D control for the sip/puff interface, and 2-D control for the joystick. Our reconstruction algorithm effectively lifts these raw demonstrations into higher simultaneous dimensional control, for all tasks and interfaces, and in some cases up to the maximum of 6-D.

Note that not all tasks require movement within all 6 control dimensions for successful execution. There are also tasks for which movement does not occur within all 6 dimensions for a given trajectory segment, as denoted by task and environment constraints. For all of these cases, the reconstructed trajectory will exist in a lower dimension than 6-D, because the (raw) demonstration for that task did not activate all 6 control dimensions, even sequentially. 

Table~\ref{tab:tally_dim_btw_t_e_constraints} reports the (maximum) number of controllable dimensions for a composed segment, averaged across demonstrations. This represents the theoretical upper bound on the number of dimensions a demonstrator might have simultaneously activated if interface constraints were absent. It is computed by tallying the dimensions activated (in sequence), during a trajectory segment demarcated by task and environment constraints (across which our algorithm will not compose). 

\begin{table} 
    \centering
    \caption{Maximum number of controllable dimensions for a composed segment,
    averaged over all demonstrations.} 
    \label{tab:interfaces_dofs}
    \resizebox{\columnwidth}{!}{
    \begin{tabular}{l|c>{\columncolor{gray!20}}cccc>{\columncolor{gray!20}}c}
        \toprule
        Interface & Pick-and-Place & Pour (1) & Peg-in-hole & Stir & Brush Hair & Cut Food (3) \\
        \midrule
        Joystick  & 4.4$\pm$0.8 & 4.8$\pm$1.0 & 4.0$\pm$0.0 & 4.0$\pm$0.0 & 4.0$\pm$0.0 & 4.0$\pm$0.0 \\
        Sip/Puff  & 3.2$\pm$0.7 & 4.0$\pm$0.0 & 2.5$\pm$0.5 & 2$\pm$0.0 & 2.7$\pm$0.9 & 3.2$\pm$0.7 \\
        \bottomrule
    \end{tabular}
    }
    \label{tab:tally_dim_btw_t_e_constraints}
    \vspace{-0.6cm}
\end{table}

Figure~\ref{fig:dof_vs_time_plot} zooms in to display
the time evolution of dimensions activated for a single demonstration trajectory (Pick-and-place task, collected via 2-D joystick). The presence of task and environment constraints are annotated.
Within the trajectory segments demarcated by these constraints, we observe that the maximum number of control dimensions activated is 6. The reconstructed algorithm, in this example, thus lifts the trajectory segment between task constraints to 6-D.

\begin{figure}[b]
    \centering
       \includegraphics[width=\linewidth]{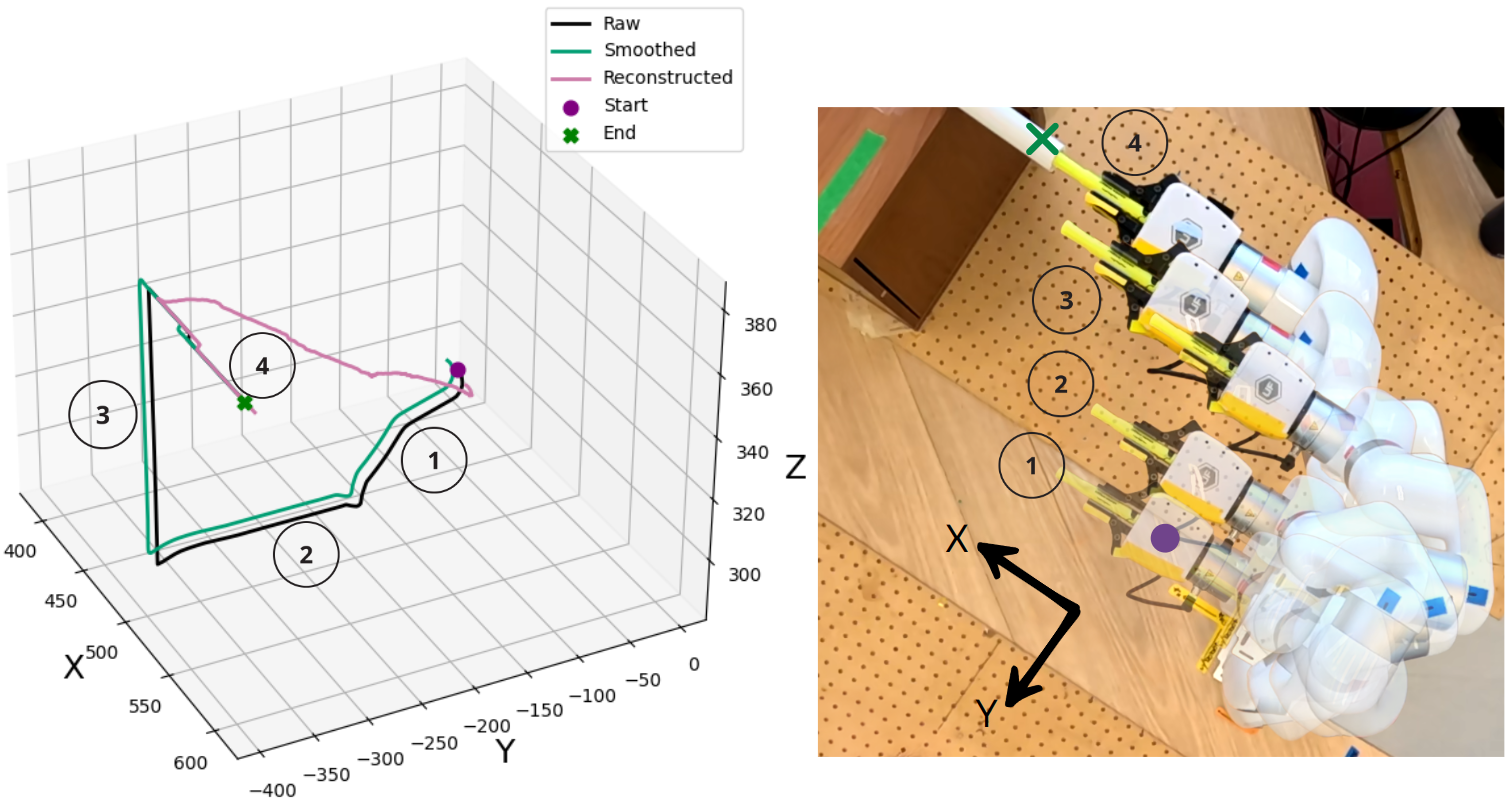}
    \caption{Comparison between interface-aware reconstruction method (magenta) and smoothing filter (green) on the Peg-in-hole raw trajectory (black, \textit{left}) and the corresponding demonstration on the 7-DoF xArm robotic arm (\textit{right}). Numbers indicate task progression.}
    \Description{This figure shows the 3-D trajectory of a Peg-in-Hole demonstration (left) and the corresponding demonstration on the xArm robotic arm (right).}
    \label{fig:filtering}
\end{figure}

\vspace{-3mm}
\paragraph{Comparison to Filtering}
To address the question of to what extent demonstrations might achieve higher dimensionality simply via smoothing, we compare three standard digital signal processing filters---Butterworth, Savitzky-Golay, and Spline (additional details in Appendix~\ref{app:filters}: Figure~\ref{fig:heatmap_smoothed}). None are able to lift a given trajectory both in a meaningful way and up more than a single dimension (Figure~\ref{fig:filtering}). 
For the joystick interface, smoothing actually decreases the dimensionality for all tasks.
Of the three, the Butterworth filter 
smooths sharp curves the most, and so is employed in the comparative analyses of the following sections. \vspace{-0.9mm}

\begin{table*}[h!]
    \centering
    \caption{Joystick interface execution time and path length, averaged over demonstrations, by and across tasks, on the xArm7. Percentages reflect changes from raw demonstrations.}
    \label{tab:joystick_all_tasks}
    \resizebox{\textwidth}{!}{
    \begin{tabular}{ll|cc c c c c|c}
    \toprule
    \textbf{Dataset} & \textbf{Metric} & \textbf{Pick-and-place} & \textbf{Pour} & \textbf{Peg-in-hole} & \textbf{Stir} & \textbf{Brush Hair} & \textbf{Cut Food} & \textbf{Average} \\
    \midrule
    \multirow{2}{*}{Raw} 
        & Time (s)      & 46.7$\pm$3.5 & 50.1$\pm$9.0 & 11.3$\pm$0.6 & 46.4$\pm$6.6 & 17.7$\pm$1.8 & 21.3$\pm$4.2 & -- \\
        & Dist. (m)  & 2.1$\pm$0.3 & 2.1$\pm$0.3 & 0.6$\pm$0.0 & 2.5$\pm$0.5 & 0.8$\pm$0.0 & 1.0$\pm$0.2 & -- \\
    \midrule
    \multirow{2}{*}{Smoothed} 
        & Time (s)      & 46.7$\pm$3.5 (0.0\%) & 50.1$\pm$9.0 (0.0\%) & 11.3$\pm$0.6 (0.0\%) & 46.4$\pm$6.6 (0.0\%) & 17.7$\pm$1.8 (0.0\%) & 21.3$\pm$4.2 (0.0\%) & 0.0\% \\ 
        & Dist. (m)  & 2.1$\pm$0.3 (-0.4\%) & 2.1$\pm$0.3 (-0.4\%) & 0.6$\pm$0.0 (-0.3\%) & 2.5$\pm$0.5 (-0.4\%) & 0.8$\pm$0.0 (-0.3\%) & 1.0$\pm$0.2 (-1.1\%) & -0.5\% \\ 
    \midrule
    \multirow{2}{*}{Recon.} 
        & Time (s)      & 30.5$\pm$2.3 (-35.0\%) & 38.3$\pm$7.3 (-23.6\%) & 9.3$\pm$0.4 (-17.4\%) & 36.7$\pm$6.9 (-21.0\%) & 13.5$\pm$1.7 (-23.9\%) & 13.9$\pm$3.6 (-34.6\%) & -25.9\% \\ 
        & Dist. (m)  & 1.7$\pm$0.2 (-16.1\%) & 1.9$\pm$0.2 (-9.0\%) & 0.5$\pm$0.0 (-13.7\%) & 2.3$\pm$0.5 (-7.6\%) & 0.8$\pm$0.0 (-9.1\%) & 0.9$\pm$0.2 (-13.2\%) & -11.5\% \\ 
    \bottomrule
    \end{tabular} }
\end{table*}

\begin{table*}[h!]
    \centering
    \caption{Sip/puff interface execution time and path length, averaged over demonstrations, by task and across tasks, on the xArm7. Percentages reflect changes from raw demonstrations. Gray-shading denotes fewer than 5 collected demonstrations, with the number of successful demonstrations collected noted in parenthesis.}
    \label{tab:sippuff_all_tasks}
    \resizebox{\textwidth}{!}{
    \begin{tabular}{ll|c>{\columncolor{gray!20}}c c c c >{\columncolor{gray!20}}c | c}
    \toprule
    \textbf{Dataset} & \textbf{Metric} & \textbf{Pick-and-place} & \textbf{Pour (1)} & \textbf{Peg-in-hole} & \textbf{Stir} & \textbf{Brush Hair} & \textbf{Cut Food (3)} & \textbf{Average} \\
    \midrule
    \multirow{2}{*}{Raw} 
        & Time (s)      & 44.5$\pm$14.0 & 37.9 & 16.8$\pm$0.3 & 24.5$\pm$1.7 & 16.3$\pm$1.8 & 18.3$\pm$2.7 & --\\
        & Dist. (m)  & 1.8$\pm$0.6 & 2.5 & 0.8$\pm$0.1 & 1.5$\pm$0.1 & 0.9$\pm$0.1 & 1.2$\pm$0.1 & --\\ 
    \midrule
    \multirow{2}{*}{Smoothed} 
        & Time (s)      & 44.5$\pm$14.0 (0.0\%) & 37.9 (0.0\%) & 16.8$\pm$0.3 (0.0\%) & 24.5$\pm$1.7 (0.0\%) & 16.3$\pm$1.8 (0.0\%) & 18.3$\pm$2.7 (0.0\%) & 0.0\% \\
        & Dist. (m)  & 1.8$\pm$0.6 (-2.1\%) & 2.4 (-3.1\%) & 0.8$\pm$0.0 (-2.5\%) & 1.5$\pm$0.1 (-2.6\%) & 0.9$\pm$0.1 (-4.9\%) & 1.1$\pm$0.1 (-3.3\%) & -3.1\% \\
    \midrule
    \multirow{2}{*}{Reconstructed} 
        & Time (s)      & 29.7$\pm$9.0 (-33.2\%) & 28.2 (-25.7\%) & 10.4$\pm$1.0 (-38.3\%) & 15.9$\pm$1.0 (-35.3\%) & 11.7$\pm$1.8 (-28.6\%) & 12.3$\pm$2.1 (-33.1\%) & -33.4\% \\
        & Dist. (m)  & 1.5$\pm$0.6 (-18.4\%) & 2.3 (-8.3\%) & 0.4$\pm$0.1 (-49.9\%) & 1.3$\pm$0.1 (-17.0\%) & 0.8$\pm$0.0 (-18.9\%) & 1.0$\pm$0.2 (-12.0\%)  & -23.6\% \\
    \bottomrule
    \end{tabular}}
\end{table*}

\subsection{Task Efficiency} 
We next evaluate the efficacy of our reconstruction algorithm with regard to metrics of task performance.
Tables~\ref{tab:joystick_all_tasks} and~\ref{tab:sippuff_all_tasks} report the execution time and distance (path length) for the raw, smoothed, and reconstructed demonstrations. Across both interfaces, all tasks, and both metrics, our interface-aware reconstruction algorithm provides measurable improvements over the raw trajectories. Specifically, the gains in execution time are on average 25\% for the joystick, and more than 30\% for the sip/puff. Distance traveled improved by more than 10\% for the joystick, and over 20\% for the sip/puff. The biggest gain we observed for a single execution was as high as 50\% (Peg-in-hole distance using the sip/puff).  

In comparison, the smoothing method (Butterworth filter) provides either no improvements or extremely modest gains. Smoothing was most effective in improving distance with the sip/puff interface (on average 3.1\%).

More generally, we observe that the impact of lifting the trajectory dimensionality scales with task complexity:
The greater the number of steps required for task completion, the more our algorithm contributes to improved reconstruction and execution.

\subsection{Learning Evaluation}
We evaluate the efficacy of our reconstruction algorithm in the context of robot learning by comparing policies trained on our reconstructed demonstrations against those trained on raw demonstrations. For each interface-task pair, we train a Multi-layer Perceptron on all but one demonstration to generate a policy, using the held-out demonstration for testing.\footnote{Note that the one task-interface pair with a single successful demonstration (Pour, sip/puff) is omitted from the learning analysis.} 

Table~\ref{tab:mse_scores} depicts the mean squared error (MSE) between the unseen demonstration and the predicted trajectory. This metric captures how well the model has learned from the available demonstration data.
Results show that reconstruction preserves learning efficiency---MSE remains comparable to raw data---while in some cases (e.g., Peg-in-hole) even improving it. 

The key takeaway is despite a limited number of demontrations, we learn successful policies (as reported in Tables~\ref{tab:learned_pickplace}--~\ref{tab:learned_peg_jaco}) using both raw and reconstructed demonstrations.
This is done without requiring more data or more complex models. This differs from common approaches that chase marginal gains through scaling data and model complexity. Our framework improves outcomes by enhancing \textit{data quality}, not by burdening the learning process.  

\begin{table}[h]
    \centering
    \caption{Mean Squared Error (MSE) quantification of errors between the demonstrations and learned trajectories. Successful policy rollouts are highlighted (in green).
    (Gray shading indicates fewer than 5 collected demonstrations. Values are unit-scaled from 0 to 1, with translation in mm and orientation in radians.) }
    \label{tab:mse_scores}
    \small
    \begin{tabular}{l|cc|cc} 
    \toprule
    \textbf{Task} &
      \multicolumn{2}{c|}{Raw} &
      \multicolumn{2}{c}{Reconstructed} \\
    \cmidrule(lr){2-3} \cmidrule(lr){4-5}
     & Joy & Sip/puff & Joy & Sip/puff \\
    \midrule
    Pick-and-Place & \cellcolor{green!20} 0.002 & \cellcolor{green!20} 0.001 & 0.007 & 0.004 \\
    Pour           & \cellcolor{green!20} 0.002 & \cellcolor{gray!40} \faTimes & 0.026 & \cellcolor{gray!40} \faTimes \\
    Peg-in-hole    & \cellcolor{green!20} 0.052 & \cellcolor{green!20} 3.776 & 0.043 & 0.661 \\
    Stir           & \cellcolor{green!20} 0.004 & \cellcolor{green!20} 9e-5 & \cellcolor{green!20} 0.002 &  \cellcolor{green!20} 2e-4\\
    Brush Hair     & \cellcolor{green!20} 0.006 & \cellcolor{green!20} 0.002 & \cellcolor{green!20} 0.002 & \cellcolor{green!20} 0.010 \\
    Cut Food      & \cellcolor{green!20} 0.001 & \cellcolor{green!40!gray!20} 0.038 & \cellcolor{green!20} 0.005 & \cellcolor{green!40!gray!20} 0.069 \\
    \bottomrule
    \end{tabular}
\end{table}


\subsubsection{Quantifying Failures}
We evaluate the learned trajectories by examining their execution outcomes. 
Table~\ref{tab:mse_scores} provides a high-level overview of task success with the execution of raw and reconstructed trained policies. (We detail the specifics of task execution for each policy rollout in Appendix~\ref{app:learning_subgoals}.) In 2 of the 6 tasks (Pick-and-Place and Pour, both involving rotation at the wrist joint), the robot's kinematics contributed to a failed rollout as joint limits were exceeded during motion. A strength of our method is the ability to generalize into a new, undemonstrated control spaces, but underlying issues with low-level control can arise
from the limitations of physical hardware. This issue presents as joint limitations when using the xArm7 platform. 

When our method is tested on a platform with all continuous joints---a 7-DoF Kinova Gen2 Jaco (Quebec, Canada)---this issue does not present. (Further details and discussion provided next, in Section~\ref{sec:jaco}.) The current iteration of our algorithm primarily focuses on the interface constraints and does not yet consider the robot's dynamics. Future work will consider how the dynamics of the robot affect the reconstructed trajectory.

For the sip/puff interface, one task---Peg-in-hole---fails due to robot over- or under-shoot near the goal (i.e., hole). Thus, while the robot does not collide or hit joint limits, it does not successfully learn to insert the peg. This may be due to difficulty with generalization,
especially when the task success tolerance is low, which might be mediated by 
one- or few-shot learning.
Similarly, for the joystick interface, the same task failed and also for reasons of precision, though in this case collision with the pipe.

\subsection{Translation to Other Hardware} \label{sec:jaco}
\begin{wrapfigure}[13]{r}{.49\linewidth}
    \centering
    \includegraphics[width=\linewidth]{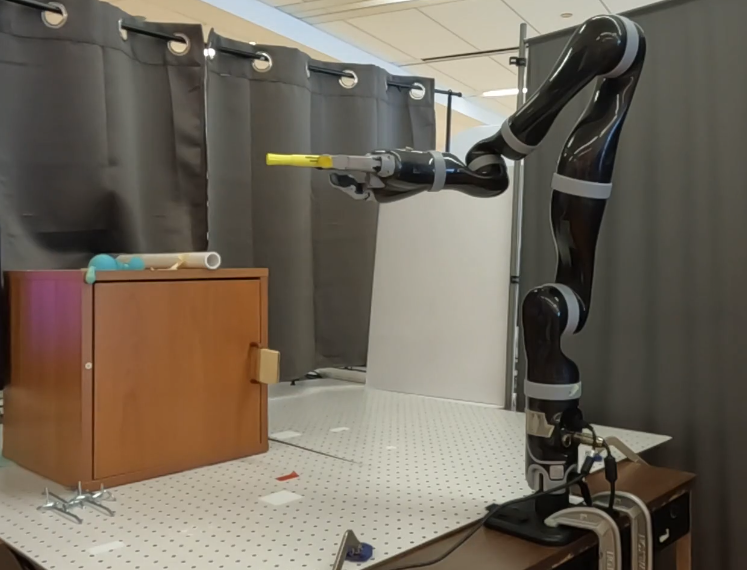}
    \caption{Peg-in-hole task setup for the Gen2 Jaco.}
    \Description{The image depicts the experimental setup for the Peg-in-hole task. The Kinova Gen2 Jaco 7-DoF robotic arm with a Sharpie highlighter in hand in its initial condition.}
    \label{fig:jaco_peghole}
\end{wrapfigure}
To demonstrate interface-aware trajectory reconstruction
on a robot with fewer physical limitations than the xArm7, we implement a subset of the tasks on a Gen2 Jaco robotic arm (Figure~\ref{fig:jaco_peghole}) from Kinova Robotics (Quebec, Canada). This 7-DoF non-anthropomorphic arm features fully continuous joints, a 3-finger gripper, and six-dimensional control (3-D translation and 3-D rotation) of the end-effector, with communication via the Kinova SDK~\cite{kinovaros}. We demonstrate on 3 of the 6 tasks: Pick-and-place, Stir\footnote{This task is a more extensive version of the Stir task for the xArm7: it additionally
includes picking up and returning the tool (whisk) for successful completion.}, and Peg-in-hole.


The Gen2 Jaco~\cite{jaco-manual} is a less dynamically constrained robotic arm compared to the xArm7~\cite{xarm-manual}, eliminating concerns about joint limits. 
The joint limit specifications for each platform are described in detail in Appendix A.4, Table~\ref{tab:joint_limits}.

Jaco success results for the sub-goals for each of the tasks are presented in detail in Appendix A.4 (Tables~\ref{tab:learned_pickplace_jaco}--\ref{tab:learned_peg_jaco}). For all tasks, including Peg-in-Hole, the robot is able to learn from both the raw and reconstructed trajectories without any hardware issues, and \emph{all} tasks succeed.

\section{Limitations and Future Work} \label{sec:limitations}
We highlight limitations of our algorithm and evaluation, and discuss future plans for improvement and expansion.

A limitation of this work is that the reported evaluations were performed by a single demonstrator who is not an end-user. While common within the robot learning literature,
within our target application domain, evaluation by end-users is paramount. Having established the ability of our algorithm to lift constrained demonstrations into higher dimensions, and its effectiveness in improving task efficiency metrics, our next step will be to engage end-users in an evaluation study. Demonstrations from end-users will provide a stress test for the robustness of the algorithm, as users, while proficient in operating their interface for wheelchair driving, likely will lack expertise in robotic arm teleoperation. The end-user study also will allow for the evaluation of policies learned on our reconstructed trajectories in terms of user preference and user feedback.

While this paper covers a range of evaluations and tasks, some tasks remain outside its scope. A more complex experimental setup, with richer representations of task and world constraints, would allow for a more extensive evaluation of the reconstruction capabilities of our algorithm.
The application of a post-processing filter also could further refine the reconstructed trajectories: 
we qualitatively observe that the reconstructed trajectories are not always smooth, 
likely due to artifacts in the reconstruction and jerkiness in the demonstrations being retained.

Our future work also will consider trajectory optimality: to identify which interface-based constraints influence the most important factors for defining optimality, and which optimality metrics to consider—such as reducing the number of mode switches, minimizing distance, or ensuring adherence to user intention.




\section{Conclusion} \label{sec:conclusion}
We have presented an interface-aware trajectory reconstruction algorithm, that lifts demonstrations constrained to lie within lower dimensions to utilize the full dimensionality of the robot control space. The reconstruction furthermore adheres to environmental and task constraints. 
For multiple tasks on two 7-DoF robot arms, teleoperated via limited 1-D sip/puff and 2-D joystick interfaces, we have confirmed that the reconstructed trajectories do
span the full control space of the robot. They also improved task performance metrics in comparison to the raw interface-constrained demonstrations.
Within the context of learning, our framework has improved task performance by enhancing data quality without increasing data or model complexity. This work establishes the foundation for teaching complex robot behaviors via demonstrations provided through limited interfaces that are accessible to end-users with motor impairments---with the potential to empower 
end-users to personalize and train their assistive robots.~\cite{xarm-manual, jaco-manual}

\begin{acks}
Funding for this work was provided by the National Science Foundation (NSF) under Grant CMMI-2208011. Any opinions, findings, and conclusions or recommendations expressed in this material are those of the authors and do not necessarily reflect the views of the NSF.

\end{acks}

\newpage
\printbibliography
\clearpage
\appendix
\section{Appendix} \label{appendix}

\subsection{Additional Filtering Details} \label{app:filters}

\subsubsection{Filtering Algorithms}
We compared the reconstructed trajectories to three commonly used forms of digital signal processing (DSP) filtering methods: (1) Low-Pass Butterworth Filter, (2) Savitzky-Golay, and (3) Spline Filtering. All filters were implemented using the open-source Python Library SciPy~\cite{scipy2020interp}. We provide the (final) hyperparameters for each method in Table~\ref{tab:filter_params}.

\begin{table}[h]
    \centering
    \caption{Filtering methods and their corresponding hyperparameters.}
    \label{tab:filter_params}
    \begin{tabular}{l|l}
        \toprule
        \textbf{Filter Type} & \textbf{Hyperparameters} \\
        \midrule
        Low-pass Butterworth & Order: 4, Cutoff: 2  \\
        Savitzky-Golay & Window length: 11, Poly order: 3  \\
        B-spline & Degree: 3  \\
        \bottomrule
    \end{tabular}
\end{table}

\begin{figure}[h]
    \centering
    \includegraphics[width=\linewidth]{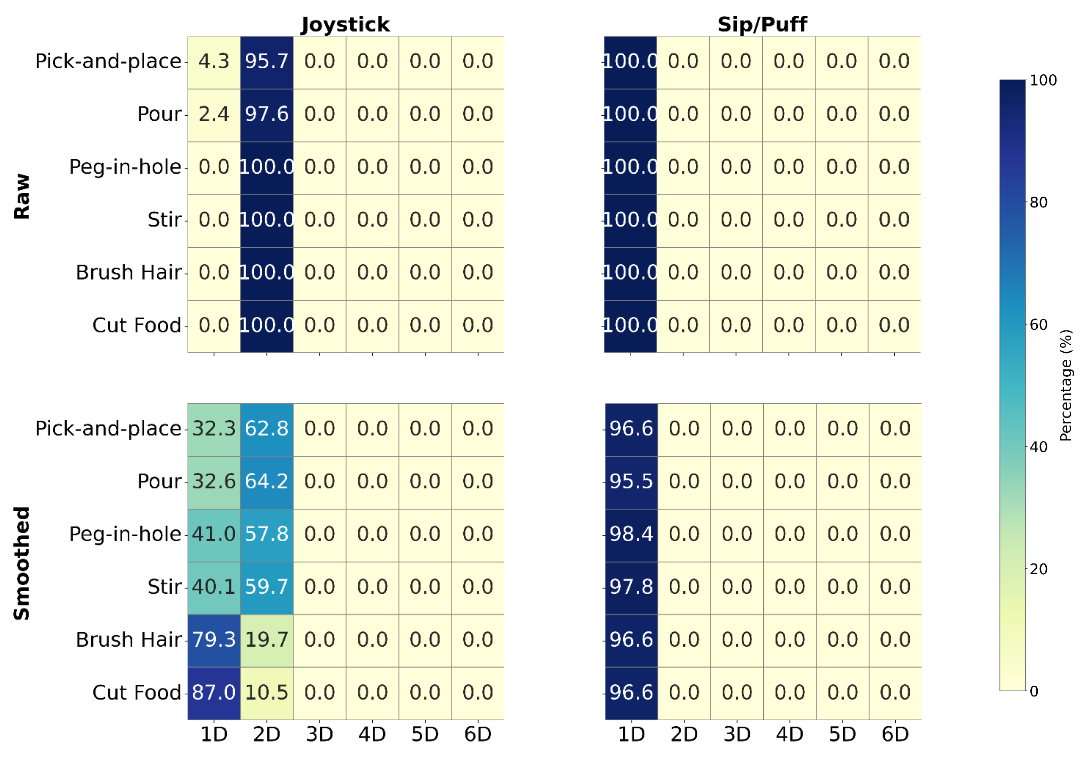}
    \caption{Percentage of dimensions activated across the full trajectory, averaged over all demonstrations. Joystick raw (\textit{top left}); joystick smoothed (\textit{bottom left}); sip/puff raw (\textit{top right}); sip/puff smoothed (\textit{bottom right}).}
    \Description{This figure includes four confusion matrices (two-by-two). The first column of matrices is for the joystick control interface while the second column of matrices is for the sip/puff control interface. The top two matrices show the percentage of dimensions activated during the raw demonstrations while the bottom two matrices show the percentage of dimensions activated after smoothing has been applied. Each row in the matrix represents one of the tasks. The joystick raw demonstrations (top left) go up to 2-D, while our reconstructed joystick demonstrations (bottom left) stays the same at 2-D. Similarly, the sip/puff raw demonstrations (top right) are only one-dimensional and same with the smoothing method (bottom right), demonstrations are up to 1-D.}
    \label{fig:heatmap_smoothed}
\end{figure}

Among the three filters evaluated, although they performed similarly, the Low-Pass Butterworth Filter performed the best at smoothing sharp curves. As a result, it was selected for the comparative analyses presented in this paper. Figure~\ref{fig:heatmap_smoothed} illustrates the percentage of activated dimensions for the raw demonstrations (top) and the smoothed demonstrations using the Butterworth Filter (bottom), with data from both the joystick interface (left) and sip/puff control (right). The heatmaps clearly show that wile the DSP filters may help reduce noise, they are unable to meaningfully enhance the dimensionality of the raw signal. In fact, the smoothing process can lead to a reduction of dimensionality. 


\subsection{Learning Algorithm Hyperparameters} \label{app:learning}

Table~\ref{tab:mlp_hyperparameters} summarizes the implementation details of the MLP trained to learn robot policies from the demonstrated data. The model architecture consisted of two hidden layers with sizes of 128 and 64 neurons, respectively, and employs the Rectified Linear Unit (ReLU) as the activation function for the hidden layers. A linear activation function is used on the output layer due to the regression nature of the predictive model. The Adam optimizer was used for model optimization, while mean squared error (MSE) served as the loss function. 
\begin{table}[h]
\centering
\caption{Hyperparameters for the Multi-Layer Perceptron (MLP) Model}
\begin{tabular}{ll}
    \toprule
    \textbf{Hyperparameter} & \textbf{Value} \\
    \midrule
    Number of Hidden Layers        & 2 \\
    Hidden Layer Sizes             & [128, 64] \\
    Activation Function            & ReLU \\
    Output Activation              & Linear \\
    Loss Function                  & MSE \\
    Optimizer                      & Adam \\
    Learning Rate                  & 0.001 \\
    Batch Size                     & 64 \\
    Number of Epochs               & [20, 100] \\
    \bottomrule
\end{tabular}
\label{tab:mlp_hyperparameters}
\end{table}

\newpage
\subsection{Additional Learning Evaluation Details} \label{app:learning_subgoals}

Tables~\ref{tab:learned_pickplace}--\ref{tab:learned_cut} in the Appendix summarize the specific details of success for each task by breaking them down into sub-goals. Evaluations are recorded by the researchers during policy rollout. Successful completion is demarcated with a checkmark (\faCheck), while unsuccessful attempts are marked with an \faTimes. Sub-goals not completed due to joint-limit violations are marked with `---'.

In two of the six tasks (Pick-and-place and Pour), the xArm7 robot's dynamics contributed to a failed rollout as joint limits were exceeded during motion. Were it not for hardware limitations, we could expect a successful rollout of the policy. (Section~\ref{sec:jaco} further elaborates on this specific hardware limitation and its implications.) 


\begin{table}[h]
    \centering
    \caption{xArm7: Sub-goal success checks for Raw and Reconstructed policies on the \textit{Pick-and-Place task}.}
    \label{tab:learned_pickplace}
    \begin{tabular}{l|cc|cc}
        \toprule
        \textbf{Sub-goal} & \multicolumn{2}{c|}{\textbf{Raw}} & \multicolumn{2}{c}{\textbf{Reconstructed}} \\ 
        \cmidrule(lr){2-3} \cmidrule(lr){4-5}
         & Joystick & Sip/Puff & Joystick & Sip/Puff \\ 
        \midrule
        Overall success?     & \faCheck & \faCheck  & \faTimes   & \faTimes  \\
        Collision free?      & \faCheck & \faCheck  & \faCheck & \faCheck \\
        Within joint limits? & \faCheck & \faCheck  & \faTimes   & \faTimes  \\
        Cup picked up?       & \faCheck & \faCheck  & \faCheck & \faCheck  \\
        Cup flipped?         & \faCheck & \faCheck  & \faCheck & \faCheck  \\
        Cup placed down?     & \faCheck & \faCheck  & --- & --- \\
        \bottomrule
    \end{tabular}
\end{table}

\begin{table}[h]
    \centering
    \caption{xArm7: Sub-goal success checks for Raw and Reconstructed policies on the \textit{Pour Task}.}
    \label{tab:learned_pour}
    \begin{tabular}{l|cc|cc}
        \toprule
        \textbf{Sub-goal} & \multicolumn{2}{c|}{\textbf{Raw}} & \multicolumn{2}{c}{\textbf{Reconstructed}} \\ 
        \cmidrule(lr){2-3} \cmidrule(lr){4-5}
         & Joystick & Sip/Puff & Joystick & Sip/Puff \\ 
        \midrule
        Overall success? & \faCheck & \faCheck & \faTimes & \faTimes \\
        Collision free? & \faCheck & \faCheck & \faCheck & \faCheck \\
        Within joint limits? & \faCheck & \faCheck & \faTimes & \faTimes \\
        Cereal picked up? & \faCheck & \faCheck & \faCheck & \faCheck \\
        Cereal in bowl? & \faCheck & \faCheck & --- & --- \\
        \bottomrule
    \end{tabular}
\end{table}

\begin{table}[h]
    \centering
    \caption{xArm7: Sub-goal success checks for Raw and Reconstructed policies on the \textit{Peg-in-hole Task}.}
    \label{tab:learned_peg}
    \begin{tabular}{l|cc|cc}
        \toprule
        \textbf{Sub-goal} & \multicolumn{2}{c|}{\textbf{Raw}} & \multicolumn{2}{c}{\textbf{Reconstructed}} \\ 
        \cmidrule(lr){2-3} \cmidrule(lr){4-5}
         & Joystick & Sip/Puff & Joystick & Sip/Puff \\ 
        \midrule
        Overall success? & \faCheck & \faCheck & \faTimes & \faTimes \\
        Collision free? & \faCheck & \faCheck & \faTimes & \faCheck \\
        Within joint limits? & \faCheck & \faCheck & \faCheck & \faCheck \\
        Peg inserted in hole? & \faCheck & \faCheck & \faTimes & \faTimes \\
        \bottomrule
    \end{tabular}
\end{table}

\begin{table}[h]
    \centering
    \caption{xArm7: Sub-goal success checks for Raw and Reconstructed policies on the \textit{Stir Task}.}
    \label{tab:learned_stir}
    \begin{tabular}{l|cc|cc}
        \toprule
        \textbf{Sub-goal} & \multicolumn{2}{c|}{\textbf{Raw}} & \multicolumn{2}{c}{\textbf{Reconstructed}} \\ 
        \cmidrule(lr){2-3} \cmidrule(lr){4-5}
         & Joystick & Sip/Puff & Joystick & Sip/Puff \\ 
        \midrule
        Overall success? & \faCheck & \faCheck & \faCheck & \faCheck \\
        Collision free? & \faCheck & \faCheck & \faCheck & \faCheck \\
        Within joint limits? & \faCheck & \faCheck & \faCheck & \faCheck \\
        Robot whisked? & \faCheck & \faCheck & \faCheck & \faCheck \\
        \bottomrule
    \end{tabular}
\end{table}

\begin{table}[h]
    \centering
    \caption{xArm7: Sub-goal success checks for Raw and Reconstructed policies on the \textit{Brush Hair Task}.} 
    \label{tab:learned_brushhair}
    \begin{tabular}{l|cc|cc}
        \toprule
        \textbf{Sub-goal} & \multicolumn{2}{c|}{\textbf{Raw}} & \multicolumn{2}{c}{\textbf{Reconstructed}} \\ 
        \cmidrule(lr){2-3} \cmidrule(lr){4-5}
         & Joystick & Sip/Puff & Joystick & Sip/Puff \\ 
        \midrule
        Overall success?       & \faCheck & \faCheck & \faCheck & \faCheck \\
        Collision free?        & \faCheck & \faCheck & \faCheck & \faCheck \\
        Within joint limits?   & \faCheck & \faCheck & \faCheck & \faCheck \\
        Brush contacted head?  & \faCheck & \faCheck & \faCheck & \faCheck \\
        Hair brushed?          & \faCheck & \faCheck & \faCheck & \faCheck \\
        \bottomrule
    \end{tabular}
\end{table}

\begin{table}[h]
    \centering
    \caption{xArm7: Sub-goal success checks for Raw and Reconstructed policies on the \textit{Cut Food Task}.}
    \label{tab:learned_cut}
    \begin{tabular}{l|cc|cc}
        \toprule
        \textbf{Sub-goal} & \multicolumn{2}{c|}{\textbf{Raw}} & \multicolumn{2}{c}{\textbf{Reconstructed}} \\ 
        \cmidrule(lr){2-3} \cmidrule(lr){4-5}
         & Joystick & Sip/Puff & Joystick & Sip/Puff \\ 
        \midrule
        Overall success? & \faCheck & \faCheck & \faCheck & \faCheck  \\
        Collision free? & \faCheck & \faCheck & \faCheck & \faCheck \\
        Within joint limits? & \faCheck & \faCheck & \faCheck & \faCheck \\
        Food cut? & \faCheck & \faCheck & \faCheck & \faCheck \\
        \bottomrule
    \end{tabular}
\end{table}



\clearpage
\subsection{Comparison to Alternative Hardware}
In this section, we provide additional analysis related to the Gen 2 Jaco, our other platform we showcase in the main text, shown in Figure~\ref{fig:jaco_peghole}.
\begin{table}[h]
    \centering
    \caption{Gen 2 Jaco: Joint limit specifications for Gen2 Jaco and xArm7~\cite{xarm-manual, jaco-manual}.}
    \resizebox{\columnwidth}{!}{%
    \begin{tabular}{c|c|c|c|c|c|c|c}
        Robot & Joint 1 & Joint 2 & Joint 3 & Joint 4 & Joint 5 & Joint 6 & Joint 7\\
        \hline
        xArm7 & $\pm$360\textdegree & -117\textdegree, ~116\textdegree & $\pm$360\textdegree & -6\textdegree, 225\textdegree & $\pm$360\textdegree & -97\textdegree,~180\textdegree & $\pm$360\textdegree \\
        \hline
        Gen2 Jaco & \multicolumn{7}{c}{$\pm27.7$ turns} \\
    \end{tabular}
    }
    \label{tab:joint_limits}
\end{table}

For the learning evaluations on the Gen2 Jaco robotic arm, we present Tables~\ref{tab:learned_pickplace_jaco}-~\ref{tab:learned_peg_jaco} detailing the sub-goals for each of the tasks. Note, that for these evaluations, using this platform, we do not need to consider the joint limits being respected as this is not an issue that arises. For all tasks, the robot is able to learn both the raw and reconstructed trajectories without any hardware limitations.

\begin{table}[h]
    \centering
    \caption{Gen 2 Jaco: Sub-goal success checks for Raw and Reconstructed policies on the \textit{Pick-and-Place task}.}
    \label{tab:learned_pickplace_jaco}
    \begin{tabular}{l|cc|cc}
        \toprule
        \textbf{Sub-goal} & \multicolumn{2}{c|}{\textbf{Raw}} & \multicolumn{2}{c}{\textbf{Reconstructed}} \\ 
        \cmidrule(lr){2-3} \cmidrule(lr){4-5}
         & Joystick & Sip/Puff & Joystick & Sip/Puff \\ 
        \midrule
        Overall success?     & \faCheck & \faCheck  & \faCheck  & \faCheck  \\
        Collision free?      & \faCheck & \faCheck  & \faCheck & \faCheck \\
        Cup picked up?       & \faCheck & \faCheck  & \faCheck & \faCheck  \\
        Cup flipped?         & \faCheck & \faCheck  & \faCheck & \faCheck  \\
        Cup placed down?     & \faCheck & \faCheck  & \faCheck & \faCheck \\
        \bottomrule
    \end{tabular}
\end{table}

\begin{table}[h]
    \centering
    \caption{Gen 2 Jaco: Sub-goal success checks for Raw and Reconstructed policies on the \textit{Stir task}. This task includes additional sub-goals.}
    \label{tab:learned_p_jaco}
    \begin{tabular}{l|cc|cc}
        \toprule
        \textbf{Sub-goal} & \multicolumn{2}{c|}{\textbf{Raw}} & \multicolumn{2}{c}{\textbf{Reconstructed}} \\ 
        \cmidrule(lr){2-3} \cmidrule(lr){4-5}
         & Joystick & Sip/Puff & Joystick & Sip/Puff \\ 
        \midrule
        Overall success?     & \faCheck & \faCheck  & \faCheck   & \faCheck  \\
        Collision free?      & \faCheck & \faCheck  & \faCheck & \faCheck \\
        
        Whisk picked up?       & \faCheck & \faCheck  & \faCheck & \faCheck  \\
        Robot whisked?         & \faCheck & \faCheck  & \faCheck & \faCheck  \\
        Whisk returned?     & \faCheck & \faCheck  & \faCheck & \faCheck \\
        \bottomrule
    \end{tabular}
\end{table}

\begin{table}[h]
    \centering
    \caption{Gen 2 Jaco: Sub-goal success checks for Raw and Reconstructed policies on the \textit{Peg-in-hole task}.}
    \label{tab:learned_peg_jaco}
    \begin{tabular}{l|cc|cc}
        \toprule
        \textbf{Sub-goal} & \multicolumn{2}{c|}{\textbf{Raw}} & \multicolumn{2}{c}{\textbf{Reconstructed}} \\ 
        \cmidrule(lr){2-3} \cmidrule(lr){4-5}
         & Joystick & Sip/Puff & Joystick & Sip/Puff \\ 
        \midrule
        Overall success?     & \faCheck & \faCheck  & \faCheck   & \faCheck  \\
        Collision free?      & \faCheck & \faCheck  & \faCheck & \faCheck \\
        
        Peg inserted?       & \faCheck & \faCheck  & \faCheck & \faCheck  \\
        \bottomrule
    \end{tabular}
\end{table}

\end{document}